\pdfoutput=1

\documentclass[11pt]{article}

\usepackage[]{acl}

\usepackage{microtype}
\usepackage[utf8]{inputenc}
\usepackage[T1]{fontenc}
\usepackage{times}
\usepackage{latexsym}
\usepackage{amsfonts}
\usepackage{amsmath}
\usepackage{amssymb}
\usepackage{amsthm}
\usepackage{relsize}
\usepackage{xspace}
\usepackage{placeins} 

\usepackage[group-separator={,}]{siunitx}

\usepackage{tikz}
\usetikzlibrary{bayesnet}
\usetikzlibrary{arrows}
\usepackage{color}
\usepackage{graphicx}
\usepackage{caption}
\usepackage{subcaption}
\usetikzlibrary{backgrounds}

\usepackage{hhline}
\usepackage{xcolor}
\usepackage{colortbl}
\definecolor{Grey}{RGB}{210,210,210}

\newcommand{\setQuote}[1]{``{#1}''}

\usepackage{algorithm}
\usepackage{algpseudocode}

\usepackage{multirow}

\makeatletter
\newcommand\footnoteref[1]{\protected@xdef\@thefnmark{\ref{#1}}\@footnotemark}
\makeatother

\newcommand\blfootnote[1]{%
  \begingroup
  \renewcommand\thefootnote{}\footnote{#1}%
  \addtocounter{footnote}{-1}%
  \endgroup
}

\usepackage{amsmath,amsfonts,bm}
\usepackage{dsfont}

\def\eqref#1{equation~\ref{#1}}

\def\1{\bm{1}}

\DeclareMathAlphabet{\mathsfit}{\encodingdefault}{\sfdefault}{m}{sl}
\SetMathAlphabet{\mathsfit}{bold}{\encodingdefault}{\sfdefault}{bx}{n}

\newcommand{\sigmoid}{\sigma}

\newcommand{\origCCS}{\textsc{origCCS}\xspace}
\newcommand{\MarginCCS}{\textsc{MarginCCR}\xspace}
\newcommand{\TripletCCS}{\textsc{TripletCCR}\xspace}
\newcommand{\OrdRegCCS}{\textsc{OrdRegCCR}\xspace}

\newcommand{\ccs}{CCS\xspace}
\newcommand{\ccr}{CCR\xspace}
\newcommand{\CCS}{Contrast-Consistent Search\xspace}
\newcommand{\CCR}{Contrast-Consistent Ranking\xspace}

\newcommand{\confidence}{\emph{confidence}\xspace}
\newcommand{\consistency}{\emph{consistency}\xspace}
\newcommand{\consistent}{\emph{consistent}\xspace}

\newcommand{\probing}{CCR probing\xspace}
\newcommand{\Probing}{CCR Probing\xspace}
\newcommand{\prompting}{prompting\xspace}

\newcommand{\SynthFacts}{\textsc{SynthFacts}\xspace}
\newcommand{\ScalarAdj}{\textsc{ScalarAdj}\xspace} 
\newcommand{\WikiLists}{\textsc{WikiLists}\xspace}
\newcommand{\SynthContext}{\textsc{SynthContext}\xspace}
\newcommand{\Reviews}{\textsc{Reviews}\xspace}
\newcommand{\EntSalience}{\textsc{EntSalience}\xspace}

\newcommand{\ItemPair}{\textsc{ItemPair}\xspace}
\newcommand{\ItemSingle}{\textsc{ItemSingle}\xspace}
\newcommand{\ItemList}{\textsc{ItemList}\xspace}

\newcommand{\ItemPairSym}{\texttt{P}\xspace}
\newcommand{\ItemSingleSym}{\texttt{S}\xspace}
\newcommand{\ItemListSym}{\texttt{L}\xspace}

\definecolor{critRGB}{RGB}{12, 18, 194}
\definecolor{itemRGB}{RGB}{133, 12, 194}
\definecolor{compRGB}{RGB}{194, 12, 133}

\definecolor{itemBcolor}{RGB}{224, 133, 2}
\definecolor{itemAcolor}{RGB}{0, 0, 153}
\definecolor{itemCcolor}{RGB}{10, 133, 1}

\newcommand{\critColor}[1]{{\color{critRGB} #1}}
\newcommand{\itemColor}[1]{{\color{itemRGB} #1}}
\newcommand{\compColor}[1]{{\color{compRGB} #1}}

\newcommand{\itemAColor}[1]{{\color{itemAcolor} #1}}
\newcommand{\itemBColor}[1]{{\color{itemBcolor} #1}}
\newcommand{\itemCColor}[1]{{\color{itemCcolor} #1}}

\newcommand{\itemPlus}{\itemAColor{x_{i}^{+}}}
\newcommand{\itemNeg}{\itemBColor{x_{i}^{-}}}

\newcommand{\itemPlusVec}{\itemAColor{\boldsymbol{x}_{i}^{+}}}
\newcommand{\itemNegVec}{\itemBColor{\boldsymbol{x}_{i}^{-}}}

\newcommand{\itemA}{\itemAColor{x_{n}^{A}}}
\newcommand{\itemB}{\itemBColor{x_{n}^{B}}}
\newcommand{\itemC}{\itemCColor{x_{n}^{C}}}

\newcommand{\itemOne}{x_{i, 1}}

\newcommand{\itemN}{x_{i, N}}

\newcommand{\itemAVec}{\itemAColor{\boldsymbol{x}_{n}^{A}}}
\newcommand{\itemBVec}{\itemBColor{\boldsymbol{x}_{n}^{B}}}

\newcommand{\probe}{f_{\theta}}
\newcommand{\probeK}{f_{\theta, k}}

\newcommand{\score}{s}
\newcommand{\scoreK}{s^{k}}

\usepackage{bbm}
\usepackage{todonotes}
\makeatletter
\newcommand*\iftodonotes{\if@todonotes@disabled\expandafter\@secondoftwo\else\expandafter\@firstoftwo\fi}  %
\makeatother

\usepackage{tipa}
\newcommand{\ethz}{\text{1}}
\newcommand{\bloom}{\text{2}}

\usepackage{cleveref}
\crefname{section}{\S}{\S\S}
\Crefname{section}{\S}{\S\S}
\crefname{table}{Tab.}{}
\crefname{figure}{Fig.}{}
\crefname{algorithm}{Algorithm}{}
\crefname{equation}{Eq.}{}
\crefname{appendix}{App.}{}
\crefname{thm}{Theorem}{}
\crefname{prop}{Proposition}{}
\crefname{cor}{Corollary}{}
\crefname{observation}{Observation}{}
\crefname{assumption}{Assumption}{}
\crefformat{section}{\S#2#1#3}

\usepackage{times}
\usepackage{latexsym}

\usepackage[T1]{fontenc}

\usepackage[utf8]{inputenc}

\usepackage{microtype}

\usepackage{inconsolata}

\title{Unsupervised Contrast-Consistent Ranking with Language Models}

\author{
Niklas Stoehr$^{\ethz}$\thanks{~ Work done during an internship at Bloomberg} \qquad Pengxiang Cheng$^{\bloom}$ \qquad Jing Wang$^{\bloom}$\\
\textbf{Daniel Preo\c{t}iuc-Pietro$^{\bloom}$} \qquad \textbf{Rajarshi Bhowmik$^{\bloom}$}
\\
$^{\ethz}$ETH Z{\"u}rich \qquad $^{\bloom}$Bloomberg\\
\footnotesize \href{mailto:niklas.stoehr@inf.ethz.ch}{\texttt{niklas.stoehr@inf.ethz.ch}} \qquad
\{{\href{mailto:pcheng134@bloomberg.net}{\texttt{pcheng134}}, \href{mailto:jwang1621@bloomberg.net}{\texttt{jwang1621}}, \footnotesize \href{mailto:dpreotiucpie@bloomberg.net}{\texttt{dpreotiucpie}}, \href{mailto:rbhowmik6@bloomberg.net}{\texttt{rbhowmik6}}\}{\texttt{@bloomberg.net}}
}
}

\begin{document}
\maketitle

\begin{abstract}
Language models contain ranking-based knowledge and are powerful solvers of in-context ranking tasks. For instance, they may have parametric knowledge about the ordering of countries by size or may be able to rank product reviews by sentiment. We compare pairwise, pointwise and listwise prompting techniques to elicit a language model's ranking knowledge. However, we find that even with careful calibration and constrained decoding, prompting-based techniques may not always be self-consistent in the rankings they produce. This motivates us to explore an alternative approach that is inspired by an unsupervised probing method called Contrast-Consistent Search (CCS). The idea is to train a probe guided by a logical constraint: a language model's representation of a statement and its negation must be mapped to contrastive true-false poles consistently across multiple statements. We hypothesize that similar constraints apply to ranking tasks where all items are related via consistent, pairwise or listwise comparisons. To this end, we extend the binary CCS method to Contrast-Consistent Ranking (CCR) by adapting existing ranking methods such as the Max-Margin Loss, Triplet Loss and an Ordinal Regression objective. Across different models and datasets, our results confirm that CCR probing performs better or, at least, on a par with prompting.

\end{abstract}



\blfootnote{\href{https://github.com/niklasstoehr/contrast-consistent-ranking}{github.com/niklasstoehr/contrast-consistent-ranking}}

\section{Introduction}

\begin{figure}[t]
 \centering
 \includegraphics[width=1.0\linewidth]{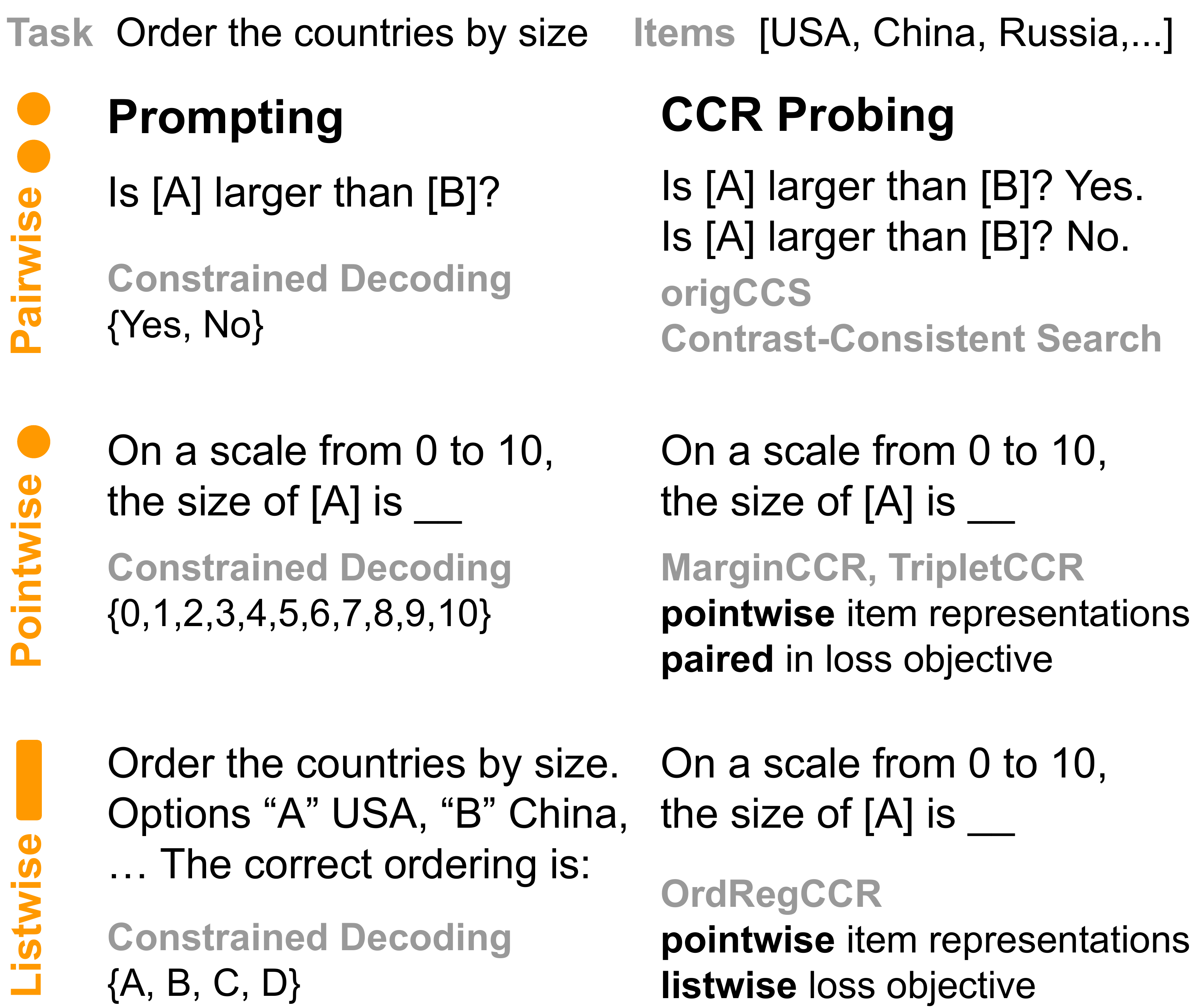} 
\caption{We study pairwise, pointwise, and listwise prompting and probing for unsupervised ranking.}
\label{fig:overview}
\end{figure}

\begin{table*}[t]
\fontsize{10}{10}\selectfont
\centering
\renewcommand{\arraystretch}{1.4} 
\setlength{\tabcolsep}{0.35em} 
\begin{tabular}{c|l|c|c}
\textbf{prompt type}       & \multicolumn{1}{c|}{\textbf{template}}                                                                                                                                                                          & \textbf{\prompting}                                                                     & \textbf{\probing}                                                                                      \\ \hline
\ItemPair \ItemPairSym     & Is \itemColor{\{item A\}} \critColor{more in terms of rank crit} than \itemColor{\{item B\}}? \compColor{$X$}                                                                                                   & \begin{tabular}[c]{@{}c@{}}constrain $X$ to\\ \compColor{\{Yes / No\}}\end{tabular}     & \begin{tabular}[c]{@{}c@{}}set $X$ to\\ \compColor{\{Yes / No\}}\end{tabular}                          \\ \hline
\ItemSingle \ItemSingleSym & \begin{tabular}[c]{@{}l@{}}\{"optional": On a scale from 0 to 10,\}\\ The \critColor{rank crit} of \itemColor{\{item\}} is \compColor{$X$}\end{tabular}                                                         & \begin{tabular}[c]{@{}c@{}}constrain $X$ to\\  \compColor{\{0, 1,...,10\}}\end{tabular} & \begin{tabular}[c]{@{}c@{}}set $X$ to\\ \compColor{[MASK]}\end{tabular}                                \\ \hline
\ItemList \ItemListSym     & \begin{tabular}[c]{@{}l@{}}\{"optional": context\}. Order by \critColor{rank crit}. Options:\\ "A" \itemColor{\{item A\}}, "B" \itemColor{\{item B\}}... .The correct ordering is: \compColor{$X$}\end{tabular} & \begin{tabular}[c]{@{}c@{}}constrain $X$ to\\ \compColor{\{A, B, ...\}}\end{tabular}    & \begin{tabular}[c]{@{}c@{}}\scriptsize embed via \ItemSingle\\  \small then listwise loss\end{tabular}
\end{tabular}
\caption{We consider three different prompt types, \ItemPair \ItemPairSym, \ItemSingle \ItemSingleSym and \ItemList \ItemListSym, that all consist of a \critColor{ranking criterion}, a \compColor{comparison token} and one or multiple \itemColor{items to be ranked}. \ItemPair and \ItemSingle can be used for \prompting and \probing in a similar fashion. To realize listwise \probing, we first obtain individual vector representations of items via \ItemSingle and then connect all items through a listwise loss objective.}
\label{tab:prompt_types}
\end{table*}

\setQuote{What is the correct ordering of the following countries by size: [USA, China, Russia, Canada, ...]?}

Language models have been shown to store plenty of facts and have powerful reasoning capacities \citep{petroni_language_2019, brown_language_2020}. Ranking tasks require both of these skills: multiple items have to be put in relation based on a comparison criterion. We are posing the question: what is the best approach to elicit a model's ranking knowledge and in-context ranking capacities without supervision? Knowing the answer to this question would allow us to uncover knowledge gaps, outdated information and existing biases before applying the language model. Once we trust a model, we could then put this best approach to action for solving in-context ranking tasks.

A natural starting point for unsupervised ranking is prompting. In \cref{sec:prompting}, we explore different task formulations: pairwise, pointwise and listwise prompting as outlined in \cref{fig:overview}. In the pairwise setting, any two items are compared and pairwise results are converted into a global ranking post-hoc. In pointwise prompting, the model assigns a score to each item individually. The listwise approach tasks the model to directly decode the entire ranking. For either approach, constrained decoding is essential to ensure the output can be converted into a ranking that includes all items. Yet, even with constrained decoding and calibration, we find that prompting often leads to inconsistent rankings.

For this reason, we turn to the model-internal representations of ranking tasks and their items in \cref{sec:probing}. We train a \setQuote{probing model} with different unsupervised ranking objectives to find a latent ordering direction in the items' vector representations. \citet{burns_discovering_2023} recently proposed the \CCS (\ccs) method to find a direction in a language model's activation space that distinguishes truthful from false statements \citep{li_inference-time_2023}. This is achieved with a loss that imposes a logical constraint: the representation of a statement and its negation must be mapped to opposite (contrastive) poles. 

Ranking tasks share similar properties: we can convert a ranking task into multiple pairwise comparisons and train a probe to find a \setQuote{ranking direction} that allows ranking one item higher than the other consistently across all pairs. This has one significant advantage over the original \ccs method for factual statements---instead of requiring a training set of multiple yes-no questions, we can source all pairwise permutations from a list of items which allows training the probe on a single ranking task.

We extend and adapt the binary \ccs method to \CCR (\ccr) by exploring pairwise (\cref{sec:pairProbing}), pointwise (\cref{sec:pointProbing}) and listwise (\cref{sec:listProbing}) approaches as illustrated in \cref{fig:overview}. Pairing items in the prompt and obtaining the vector representations of all pairs is computationally expensive. Moreover, binary, contrastive poles may not be ideally suited for ranking tasks where the distances between items are not unit-length. Similar to the pointwise prompting approach, we instead embed each item individually, e.g., \setQuote{The size of the US is [MASK], The size of China is [MASK], ...}. We then pair the items represented by the activations of the [MASK] tokens in the loss function. In particular, we propose variants of the well-known Max-Margin and Triplet loss by including a \consistency and \confidence component. As a final adjustment, we mitigate the limitation that pairwise and pointwise objectives do not guarantee transitivity: item A may be ranked above B, B above C, but C above A, creating a circular contradiction. To address this, we introduce an unsupervised ordinal regression objective for listwise \probing. 

Our experiments in \cref{sec:experiments} confirm that \probing outperforms \prompting with a DeBERTa \citep{he_deberta_2021} and GPT-2 \citep{jiang_how_2021} model across six datasets. Among the \probing methods, the Triplet Loss variant performs best on average. Even for a much larger MPT-7B \citep{mosaicml_mpt-7b_2023} model, \probing performs at least on a par with \prompting. Yet, \probing has the advantage of better control, reliability and interpretability as we discuss in \cref{sec:discussion}.

\section{Prompting for Rankings}
\label{sec:prompting}

Prompting is an accessible way to test a language model's ranking knowledge \citep{li_probing_2022}. We experiment with three different prompt types outlined in \cref{tab:prompt_types}: pairwise, pointwise and listwise prompting \citep{qin_large_2023}. All prompt types contain at least one \itemColor{item to be ranked}, a \critColor{criterion to rank on}, and what we refer to as \compColor{comparison token}. In every setting, we rely on some form of \setQuote{constrained decoding} (for decoder-only) or \setQuote{constrained mask-filling} (for encoder-only models). In essence, we restrict the vocabulary to a list of candidates and select the tokens with the highest-scoring logits.

\paragraph{Pairwise Prompting.}

\ItemPair \ItemPairSym:  \emph{Is \itemColor{\{item A\}} \critColor{more in terms of ranking criterion} than \itemColor{\{item B\}}? \compColor{Yes / No}}---Between any two items, the language model is tasked to make ranking decisions which are then converted into a ranking post-hoc as elaborated on in \cref{sec:evaluation}. Without calibration \citep{zhao_calibrate_2021}, the model tends to always output the token most frequently observed during training, disregarding the task. Following \citep{burns_discovering_2023}, we compute the mean logit score of the \setQuote{Yes} and \setQuote{No} tokens in all pairwise prompts and then subtract the respective mean from each token's score. 

\paragraph{Pointwise Prompting.}

\ItemSingle \ItemSingleSym:  \emph{On a scale from \num{0} to \num{10}, the \critColor{ranking criterion} of \itemColor{\{item\}} is \compColor{$X$}}---In pointwise prompting, the language model ranks one item at a time. If two items are assigned the same rank (i.e., the same candidate token from the list $\compColor{X} \in \{0,1,2,\ldots,10\}$), we break the tie via sorting by the tokens' logit scores.

\paragraph{Listwise Prompting.}

\ItemList \ItemListSym  \emph{optional: context. Order by \critColor{ranking criterion}. Options: \setQuote{A} \itemColor{\{item A\}}, \setQuote{B} \itemColor{\{item B\}}... The correct ordering is: \compColor{$X$}}---For listwise prompting, we apply a step-wise approach: we let the model select the highest-scoring item from the list of candidates $\compColor{X} \in \{A, B, ...\}$, remove this token from the list and append it to the prompt. We repeat the process until the candidate list is exhausted. Importantly, the ordering of the candidate options in the prompt poses a \setQuote{positional bias} \citep{han_prototypical_2023, wang_large_2023}. Therefore, we randomly shuffle the ordering of the options and repeat the listwise prompting multiple times.

\begin{table}[t]
\fontsize{9}{9}\selectfont
\centering
\renewcommand{\arraystretch}{1.3} 
\setlength{\tabcolsep}{0.20em} 
\begin{tabular}{c|cc|cc}
                                            & \textbf{prompt type}       & \textbf{emb calls}   & \textbf{loss / model} & \textbf{datapoints}  \\ \hline
\multirow{5}{*}{\rotatebox{90}{\probing}}   & \ItemPair \ItemPairSym     & $\mathcal{O}(N^{2})$ & $\origCCS$            & $\mathcal{O}(N^{2})$ \\
                                            & \ItemSingle \ItemSingleSym & $\mathcal{O}(N)$     & $\origCCS$            & $\mathcal{O}(N^{2})$ \\
                                            & \ItemSingle \ItemSingleSym & $\mathcal{O}(N)$     & $\MarginCCS$          & $\mathcal{O}(N^{2})$ \\
                                            & \ItemSingle \ItemSingleSym & $\mathcal{O}(N)$     & $\TripletCCS$         & $\mathcal{O}(N^{3})$ \\
                                            & \ItemSingle \ItemSingleSym & $\mathcal{O}(N)$     & $\OrdRegCCS$          & $\mathcal{O}(N)$     \\ \hline
\multirow{3}{*}{\rotatebox{90}{\prompting}} & \ItemPair \ItemPairSym     & $\mathcal{O}(N^{2})$ & MLM / causal          & $\mathcal{O}(N)$     \\
                                            & \ItemSingle \ItemSingleSym  & $\mathcal{O}(N)$     & MLM / causal          & $\mathcal{O}(N)$     \\
                                            & \ItemList \ItemListSym     & $\mathcal{O}(N)$     & MLM / causal          & $\mathcal{O}(1)$    
\end{tabular}
\caption{Complexity of each approach as a factor of the number of items $N$ per ranking task. We distinguish between the number of required calls of an \setQuote{embedding function} (i.e., a language model) and the number of resulting data points to be considered in a subsequent loss objective. The asymptotic complexity of permutations and combinations is both $\mathcal{O}(N^{2})$.}
\label{tab:complexity}
\end{table}

\section{Unsupervised Probing for Rankings}
\label{sec:probing}

Querying a language model's knowledge via prompting, we limit ourselves to prompt design and evaluating the tokens' logit scores. In contrast, probing accesses the information contained within a language model more directly by operating on latent vector representations. Conventionally, probing involves training a \setQuote{diagnostic classifier} to map the vector representations of an utterance to a target label of interest (e.g., tense, gender bias, etc.) in a supervised fashion. The goal typically is to measure what information is contained within a language model \citep[\emph{inter alia}]{alain_understanding_2016, belinkov_what_2017}. While the motivation of this work is closely related, we focus on an unsupervised probing variant and consider supervised probing only as a performance upper bound for validation purposes in \cref{sec:results} and \cref{sec:discussion}.

\paragraph{\CCS (\ccs).}

\citet{burns_discovering_2023} propose \CCS (\ccs), an unsupervised probing method which seeks to train a probe to satisfy logical constrains on the model's activations. Instead of labels, \ccs requires paired prompts in the form of yes-no questions:
\begin{align}
\label{eq:binaryPrompt}
\itemPlus &= \text{\setQuote{Are elephants mammals? \itemAColor{Yes}}}\\
\itemNeg &= \text{\setQuote{Are elephants mammals? \itemBColor{No}}} \nonumber
\end{align}
Both statements $\itemPlus$ and $\itemNeg$ are fed to a language model and the activations of the model's last hidden layer corresponding to the \setQuote{Yes} and \setQuote{No} token, $\itemPlusVec$ and $\itemNegVec$ (bolded), are considered in subsequent steps. First, the vector representations $\itemPlusVec$ and $\itemNegVec$ from different yes-no questions have to be Z-score normalized to ensure they are no longer forming two distinct clusters of all \setQuote{Yes} and \setQuote{No} tokens. Next, the paired vectors are projected to a score value $\score_{i}$ via the probe $\probe(\boldsymbol{x}_{i}) = \sigmoid(\boldsymbol{\theta}^{T} \boldsymbol{x}_{i} + b)$ which is trained using the \origCCS loss objective:
\begin{align}
\origCCS = &\overbrace{\Big(\probe (\itemPlusVec) - \big(1 - \probe (\itemNegVec)\big)\Big)^{2}}^\text{\consistency}\label{eq:origCCS}\\ 
&+ \underbrace{\min\big(\probe (\itemPlusVec), \probe (\itemNegVec)\big)^{2}}_\text{\confidence}\nonumber
\end{align}
\origCCS comprises two terms: the \consistency term encourages $\probe (\itemPlusVec)$ and $\probe (\itemNegVec)$ to sum up to \num{1}. The \confidence term pushes the scalars away from a deficient $\probe(\itemPlusVec) = \probe(\itemNegVec) = 0.5$ solution, and instead encourages one to be close to \num{0} and the other to be close to \num{1}. Thus, the \origCCS objective promotes mapping true and false statements to either \num{0} or \num{1} consistently, when the probe is trained on multiple yes-no questions.\footnote{\ccs (and \ccr) are direction-invariant, see \cref{sec:ranking_direction}.}  

\paragraph{From Yes-No Questions to Rankings.}

\origCCS relies on logical constraints to identify a true-false mapping in the models' activations. We argue that ranking properties can similarly be expressed as logical constraints which are discernable by a probing model. In fact, the pairing of yes-no statements in \cref{eq:binaryPrompt} resembles the \ItemPair prompt type presented in \cref{tab:prompt_types}.

One advantage of ranking tasks is that we can source many pairwise comparisons from a single ranking task which reduces the need for a training set of different yes-no questions. In the original \ccs paper, it has been shown that a training set of as few as \num{8} pairwise comparisons can be enough for good test set performance. A ranking task of eight items allows for \num{28} comparisons when considering all pairwise combinations and even \num{56} comparisons when considering all pairwise permutations.

We adapt binary \ccs to \CCR (\ccr) by gradually modifying three components of the original method: in \cref{sec:pairProbing}, we start by changing only the prompt. In \cref{sec:pointProbing}, we explore pointwise \probing which requires changing the loss function in addition. Finally, in \cref{sec:listProbing}, we also alter the probing model to propose a listwise regression approach. Importantly, all \ccr approaches are unsupervised and involve training a linear probing model whose number of parameters is held constant across settings to allow for a fair comparison. 

\subsection{Pairwise \Probing}
\label{sec:pairProbing}

Pairwise \probing for rankings is straight-forward as we only need to change the binary prompt in \cref{eq:binaryPrompt} to the \ItemPair \ItemPairSym prompt type in \cref{sec:pairProbing}, but apply the original \origCCS objective (\cref{eq:origCCS}), which we abbreviate as \setQuote{\origCCS (\ItemPairSym)}.

\subsection{Pointwise \Probing}
\label{sec:pointProbing}

We observe several methodological shortcomings of the pairwise \probing approach based on \origCCS that we address in the following. We start with the observation that it is computationally expensive to \setQuote{embed} all pairwise item permutations as depicted in \cref{tab:complexity}. Instead, we propose to \setQuote{embed} each item individually and to pair their representations in the subsequent loss objective. To this end, we consider the \ItemSingle \ItemSingleSym prompt type for \probing which requires much fewer \setQuote{calls} of a language model, precisely as many as there are items in a ranking task:
%
\begin{align}
\label{eq:ItemSingle}
\itemOne &= \text{\setQuote{The size of \{country 1\} is [MASK]}} \nonumber\\
\ldots\\
\itemN &= \text{\setQuote{The size of \{country N\} is [MASK]}} \nonumber
\end{align}
In the original \ccs approach, one data point $i$ is given by a binary yes-no question. Adapted to ranking, we denote a ranking task with $i$ and index its $N$ items with $n$. Since we never compare items between different ranking tasks, we omit the $i$ index for simplicity in the following. Now, the probing model $\probe$ assigns a ranking score $\score_{n} = \sigmoid(\boldsymbol{\theta}^{T} \boldsymbol{x}_{n} + b)$ directly to each item $x_n$. Scores $\score_{n}$ can then be directly plugged into the \origCCS objective, instead of $\probe(\boldsymbol{x}_{i})$, resulting in \setQuote{\origCCS (\ItemSingleSym)}.

However, the \origCCS loss enforces a hard binary decision, while an important property of rankings is that the distances between items do not necessarily have unit length. This \setQuote{ordinal property} is typically reflected by some notion of \setQuote{margin} in existing ranking objectives such as the Max-Margin and Triplet loss. To incorporate this, we propose the \MarginCCS loss which represents a modification of the well-known Max-Margin loss. 
\begin{align}
\min \bigg( &\max \Big( 0, \big(\probe(\itemAVec)-\probe(\itemBVec)\big) + m\Big), \label{eq:marginCCS}\\
 &\max \Big( 0, \big(\probe(\itemBVec)-\probe(\itemAVec)\big) + m\Big) \bigg) \nonumber
\end{align}
\MarginCCS enforces that $\itemA$ ranks higher or lower than $\itemB$ by at least a margin $m$ which can be seen as a \confidence property. Since there are no labels however, the probe has to figure out whether scoring $\itemA$ higher or lower than $\itemB$ yields better \consistency and reduces the loss across all item pair permutations.

In a similar style, we can adapt the popular Triplet Loss to \TripletCCS. To simplify notation, we denote the distance $|\probe(\itemAVec) - \probe(\itemBVec)|$ between two items $\itemA$ and $\itemB$ as $d(\itemA, \itemB)$ and compute \TripletCCS according to:
\begin{align}
\min \Big(&\max \big( 0, d(\itemC, \itemA)-d(\itemC, \itemB) + m\big), \nonumber\\
&\max \big( 0, d(\itemC, \itemB))-d(\itemC, \itemA) + m\big) \Big) \nonumber
\end{align}
Intuitively, the objective forces the \setQuote{positive item} to be closer to a third item $\itemC$, referred to as \setQuote{anchor}, than a \setQuote{negative item}, plus a \confidence margin $m$. Yet, this is enforced without knowing which item is to be labeled as \setQuote{positive} and \setQuote{negative}. Instead, the probe is trained to make this decision by being \consistent across all items in a given ranking task. We refer to both presented methods as \setQuote{\MarginCCS (\ItemSingleSym)} and \setQuote{\TripletCCS (\ItemSingleSym)} and provide further technical details on batching and vector normalization in \cref{sec:technical_details}.

\subsection{Listwise \Probing}
\label{sec:listProbing}

\begin{figure}[t]
 \centering
 \includegraphics[width=1.0\linewidth]{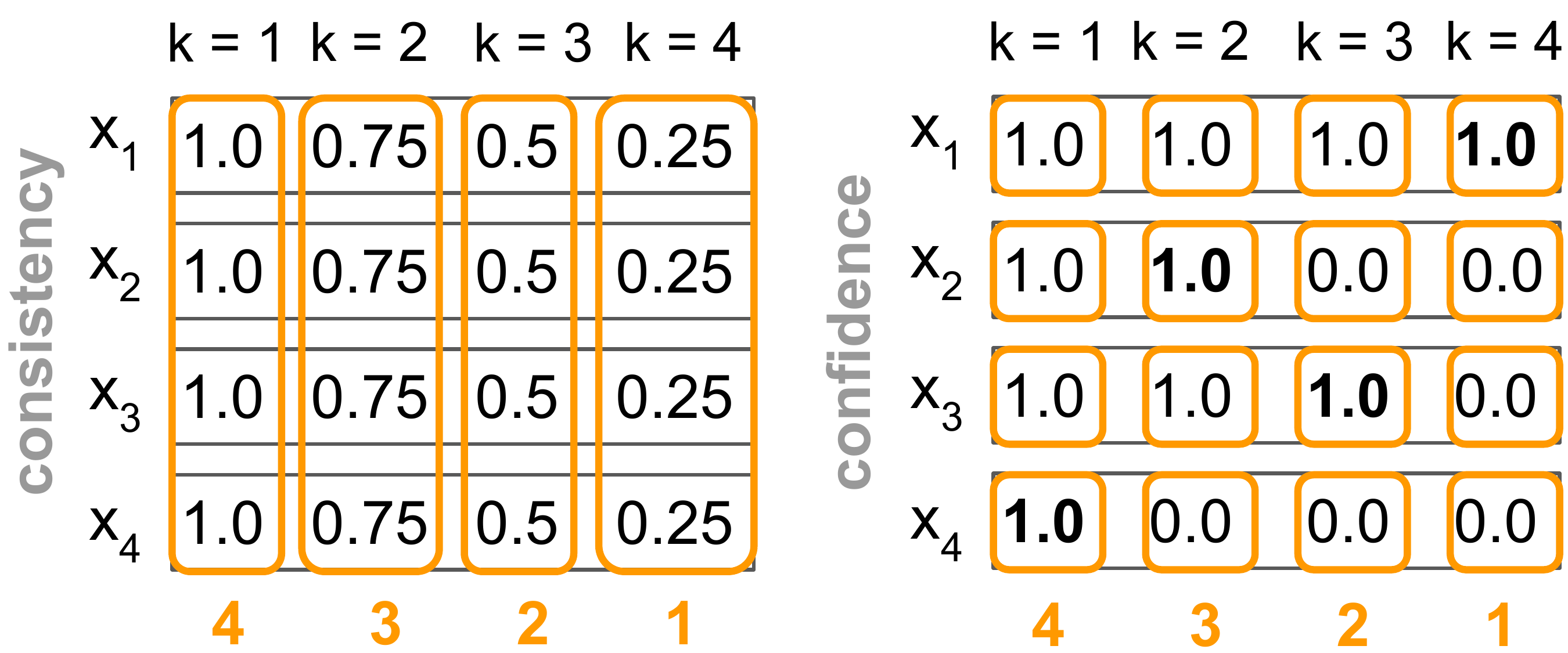} 
\caption{We translate the two aspects of \consistency and \confidence from the binary \ccs objective to an ordinal multi-class setting resulting in \OrdRegCCS.}
\label{fig:ordRegCCS}
\end{figure}

Pairwise methods are not guaranteed to yield transitivity-consistent rankings: item A may win over B, B may win over C, yet C may win over A, creating a circular ordering \citep{cao_learning_2007}. To tackle this shortcoming, we design a listwise probing method with a loss objective that considers all items at the same time. Various existing ordinal regression methods are based on binary classifiers \citep{li_ordinal_2006, niu_ordinal_2016, shi_deep_2021}, making them a natural candidate for a \ccs-style objective that does not require more parameters. These methods often rely on the extended binary representation \citep{li_ordinal_2006} of ordered classes, where, for instance, rank $k=3$ out of $K=4$ would be represented as $[1,1,1,0]$, as illustrated on the right side of \cref{fig:ordRegCCS}.

We first obtain a vector representation $\boldsymbol{x}_n$ of item $x_n$ using the \ItemSingle prompt type. Next, we consider the COnsistent Rank Logits (CORAL) model \citep{cao_rank_2020}, which offers guarantees for rank-monotonicity by training a probe $\probeK$ to map $\boldsymbol{x}_n$ to one of $K$ ranks. The probe consists of the weight vector $\boldsymbol{\theta}^{T}$ and $K$ separate bias terms $b_{k}$ to assign a rank score $\scoreK_{n}$ according to $\scoreK_{n} = \probeK(\boldsymbol{x}_n) = \sigmoid(\boldsymbol{\theta}^{T} \boldsymbol{x}_n + b_{k})$. In essence, for each item $n$, the CORAL probe outputs a vector of $K$ scores. Scores are monotonically decreasing because the bias terms $b_{k}$ are clipped to be monotonically decreasing as $k$ grows larger. Predicting a rank in the extended binary representation thus comes down to $\hat{k} = 1 + \sum_{k=1}^{K} \mathds{1}[ \scoreK > 0.5]$.

In a listwise approach, all $N$ items are to be jointly considered and assigned a rank $k$.\footnote{We note that the number of ranks $K$ equals the number of items $N$, but keep both letters for notational simplicity.} The predicted scores can thus be represented as a square $N \times K$ matrix as displayed in \cref{fig:ordRegCCS}. We propose an unsupervised ordinal regression objective that encourages a unique rank assignment, which we term \OrdRegCCS:
\begin{align}
\overbrace{\sum_k^{K-1} \Big( \big(K- (k-1) \big) -  \sum_{n}^{N} \scoreK_{n} \Big)}^\text{\consistency} +
\nonumber\\
\underbrace{\sum_n^{N} \sum_k^{K} \min \Big(\scoreK_{n},  \big( 1-\scoreK_{n} \big) \Big)}_\text{\confidence} \label{eq:ordRegCCS}
\end{align}
For a ranking of $K=4$ items, the \consistency term encourages each column to sum up to \num{4}, \num{3},..., \num{1} respectively, as visualized in \cref{fig:ordRegCCS}. Yet, to avoid a degenerate solution, the \confidence term enforces each score towards either \num{0} or \num{1}.

When applying this \setQuote{\OrdRegCCS (\ItemSingleSym)} approach, there are two difficulties to overcome: firstly, we require the number of parameters of the probing model to be the same across different approaches to ensure a fair comparison. Secondly, we prefer training a probing model whose parameters are independent from the number of items of a given ranking task. To mitigate both issues, we parametrize the $K$ bias terms via a polynomial function as elaborated in \cref{sec:bias_terms}. This function, in turn, is parametrized by only two parameters, $\alpha$ and $\beta$, which are optimized during training.

\begin{table*}[t]
\fontsize{10}{10}\selectfont
\centering
\renewcommand{\arraystretch}{1.2} 
\setlength{\tabcolsep}{0.50em} 
\begin{tabular}{l|lccl}
                                            & \textbf{dataset} & \textbf{tasks} & \textbf{avg. items} & \textbf{ranking example}                                                                                                                                                                                                         \\ \hline \hline
\multirow{5}{*}{\rotatebox{90}{fact-based}} & \SynthFacts      & 2              & 6.00           & \begin{tabular}[c]{@{}l@{}}criterion: order the numbers by cardinality\\ items: \{1, 10, 100, 1000...\}\end{tabular}                                                                                                             \\ \cline{2-5} 
                                            & \ScalarAdj       & 38             & 4.47           & \begin{tabular}[c]{@{}l@{}}criterion: order the adjectives by semantic intensity\\ items: \{small, smaller, tiny, microscopic...\}\end{tabular}                                                                                  \\ \cline{2-5} 
                                            & \WikiLists       & 69             & 9.23          & \begin{tabular}[c]{@{}l@{}}criterion: order the countries by size\\ items: \{Russia, Canada, China, United States...\}\end{tabular}                                                                                              \\ \hline\hline
\multirow{7}{*}{\rotatebox{90}{context-based}} & \SynthContext    & 2              & 6.00           & \begin{tabular}[c]{@{}l@{}}context: ``Tom owns \$100, Jenny has \$1000,...''\\ items: \{Tom, Jenny, Emily, Sam...\}\\ criterion: order entities by wealth\end{tabular}                                                           \\ \cline{2-5} 
                                            & \Reviews         & 805             & 5.00           & \begin{tabular}[c]{@{}l@{}}context: \{A: I endorse this product..., B: The product is bad...\}\\ items: \{review A, review B...\}\\ criterion: order the product reviews by stance\end{tabular}                                  \\ \cline{2-5} 
                                            & \EntSalience     & 362            & 7.50            & \begin{tabular}[c]{@{}l@{}}context: ``The UN secretary met with climate activists...''\\ items: \{UN secretary, climate activists, US government...\}\\ criterion: order the entities by salience in the given text\end{tabular}
\end{tabular}
\caption{Overview of datasets, their number of ranking tasks and the average number of items per task. The first three datasets require knowledge of facts (fact-based), the latter three require in-context reasoning (context-based).}
\label{tab:data}
\end{table*}

\begin{figure*}[t]
 \centering
 \includegraphics[width=1.0\linewidth]{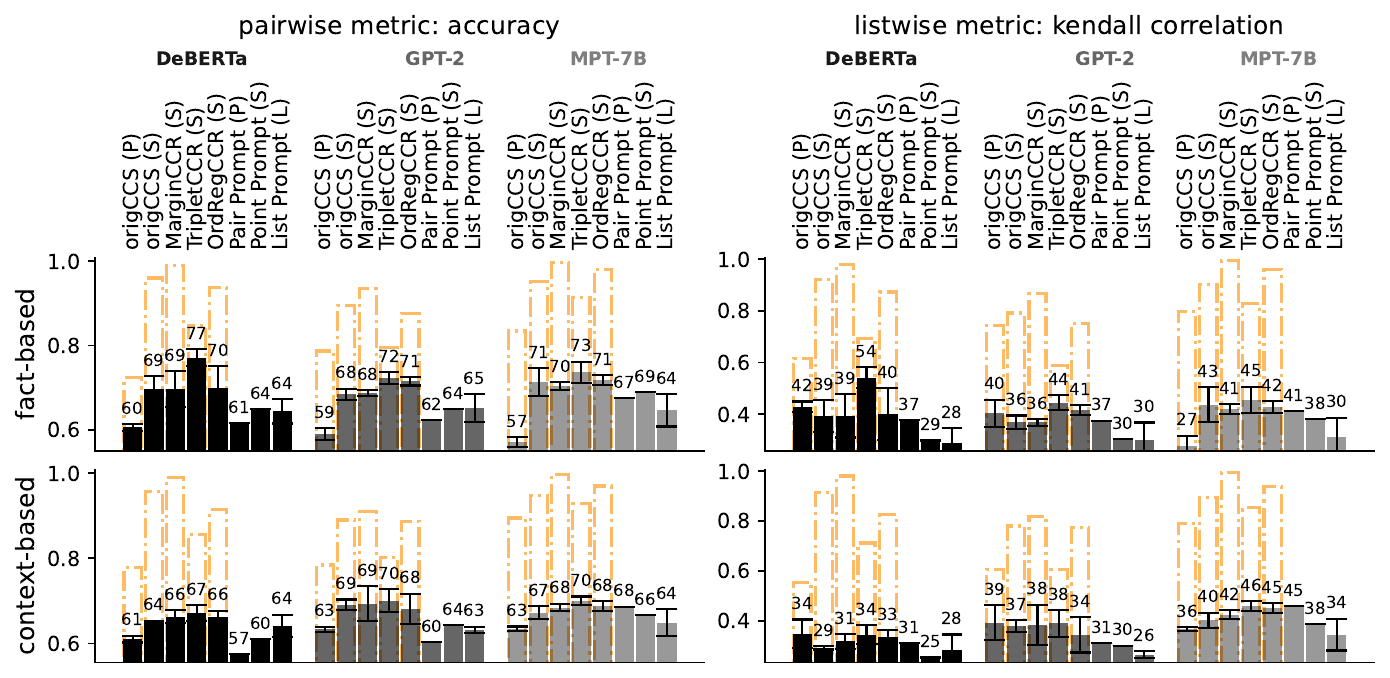} 
\caption{Pairwise and listwise results of the \prompting and \probing methods for the DeBERTa, GPT-2 and MPT-7B model, meaned over all fact-based and context-based learning datasets. Results show mean and standard deviation over \num{5} runs. We find that \probing often outperforms \prompting for the same-size model. Among the \probing methods, \TripletCCS is the best-performing. Orange bars represent ceilings of a supervised probe trained and tested on the same ranking task. As model size increases (MPT-7B), prompting performance improves.}
\label{fig:per_ranking_mean}
\end{figure*}

\section{Experimental Design}
\label{sec:experiments}

\subsection{Language Models}

We evaluate the \prompting and \probing methods on an encoder-only and a decoder-only model. For the encoder-only model, we choose \href{https://huggingface.co/microsoft/deberta-base}{\texttt{deberta-v1-base}} \citep{he_deberta_2021} which has \num{100} million parameters and is the best-performing encoder-only model for answering yes-no questions in the original \ccs paper. For the decoder-only model, we consider \href{https://huggingface.co/gpt2}{\texttt{gpt2}} (small) \citep{jiang_how_2021} which has \num{124} million parameters. We compare these models against prompting results achieved with a much bigger, \num{7} billion parameter \href{https://huggingface.co/mosaicml/mpt-7b}{\texttt{mpt-7b}} \citep{mosaicml_mpt-7b_2023} model. 

\subsection{Ranking Task Datasets}

We consider two types of ranking tasks with three datasets each. We denote the first type ``fact-based'' as solving it depends mostly on world knowledge. In contrast, the required information for the second type is provided ``in-context''. All datasets, displayed in \cref{tab:data}, are publicly available and we discard all ranking tasks with fewer than four items and those including ties between items.

\paragraph{Fact-based Ranking Tasks.}

\SynthFacts: we manually conceive two synthetic ranking tasks with six items each. One task asks to rank adjectives based on sentiment, the other to rank numbers based on cardinality (App. \cref{tab:synth_data_details}). \ScalarAdj: we consider rankings of \href{https://github.com/ainagari/scalar_adjs/tree/master/data}{scalar adjectives} based on \citet{de_melo_good_2013} and curated by \citet{gari_soler_bert_2020}, which are ordered by their semantic intensity, e.g., \setQuote{small, smaller, tiny,...}. \WikiLists: we manually curate a dataset of \num{69} ranking tasks based on constant or rarely changing facts about the world and cap each task at \num{10} items at maximum (see App. \cref{tab:wiki_data}).

\paragraph{In-Context Ranking Tasks.}

\SynthContext: analogously to \SynthFacts, we design two synthetic in-context ranking tasks (App. \cref{tab:synth_data_details}). The first concerns ranking colors by popularity where the popularity is unambiguously stated in a prepended context. The second task is to order entities by their wealth as described in a prepended context. \Reviews: We consider reviews and their ratings pertaining to the same product / company from the \href{https://bitbucket.org/lowlands/release/src/master/WWW2015/}{TrustPilot} dataset \citep{hovy_user_2015}, particularly the US geo-coded version. \EntSalience: As another in-context ranking task, we consider the Salient Entity Linking task (SEL) \citep{trani_sel_2016}. Given a news passage, we ask the model to rank the mentioned entities by salience.

\subsection{Evaluation Metrics}
\label{sec:evaluation}

We are considering pairwise, pointwise and listwise approaches as displayed in \cref{tab:prompt_types}. This means, we need to convert pairwise results to a listwise ranking and vice versa and consider evaluation metrics for pairwise as well as listwise results. Following the original CCS method, our evaluation is direction-invariant as further discussed in \cref{sec:ranking_direction}. In essence, the ranking $A > B > C$ is considered the same as $C > B > A$.

\paragraph{Pairwise Metric and Conversion to Ranking.}

We rely on accuracy to evaluate pairwise comparisons. To account for direction-invariance, we reverse the predicted order if the reverse order yields better results. This means that accuracy will always be $\geq 50\%$. For aggregating pairwise results into a listwise ranking, we follow \citet{qin_large_2023}: if an item wins a pairwise comparison it gains a point and points are summed to obtain a ranking. If the sum of wins is tied between items, we break the tie by considering the sum of the items' logit scores for all comparisons. 

\paragraph{Ranking Metric and Conversion to Pairs.}

To evaluate rankings, we consider Kendall's tau correlation which is independent of the number of items per ranking task and the directionality of the ordering. These desiderata are not given by other ranking and retrieval metrics such as the Normalized Discounted Cumulative Gain (NDCG) \citep{wang_theoretical_2013}. A Kendall's tau of \num{0} represents the baseline of \setQuote{no correlation} while \num{1} indicates an entirely correct ordering. We derive pairwise comparisons from a ranking by simply permuting and labeling any two items.

\subsection{Supervised Ceilings}
\label{sec:ceiling}

Both the \prompting as well as \probing approaches can be applied in an unsupervised way, thus not requiring a train-test split. We also consider a supervised probe to obtain a performance upper bound that offers an indication on the difficulty of a task and the suitability of a certain prompt design. For instance, if a prompt is entirely random, even a supervised probe would not be able to discriminate between different items. For the supervised probe, we rely on the unaltered, original loss functions, e.g., Binary Cross-Entropy instead of \origCCS, Max-Margin loss instead of \MarginCCS, etc. (see \cref{fig:k-fold} for an overview). Importantly, in \cref{sec:results}, we do not consider a train-test split and thus train and test the supervised probe on the same ranking task. In \cref{sec:discussion}, we consider a more traditional setting, where we train the probing model on ranking tasks that are distinct from the ones that we test it on.

\subsection{Results}
\label{sec:results}  

We present the results averaged over all datasets containing either fact-based or context-based ranking tasks in \cref{fig:per_ranking_mean}. All individual results are provided in \cref{fig:per_ranking} in the appendix. Most importantly, we find that \probing outperforms \prompting for the smaller-size models, DeBERTa and GPT-2. For the much larger MPT-7B model, \probing and \prompting yield narrower gaps in performance, potentially because of the stronger reasoning capabilities that boost the prompting performance of the larger models \citep{amini_probing_2023}. Among the \probing methods, \TripletCCS is the best performing approach across all models and datasets. The orange dashed lines represent the supervised ceilings for each of the \probing approaches as motivated in \cref{sec:ceiling}. Between the fact-based and context-based datasets, performance drops overall, but more for the encoder-only DeBERTa model. When considering the listwise metric, our results confirm that listwise prompting is inferior to pairwise and surprisingly also to pointwise prompting \citep{qin_large_2023, liusie_zero-shot_2023}. However, pairwise methods, here indicated with a \ItemPairSym symbol, are also computationally more expensive, making \probing even more favorable. For pairwise methods, we observe a bigger discrepancy between the pairwise accuracy and listwise kendall correlation metric. This stems from the fact that pairwise methods are more fault-tolerant---some of the pairwise comparisons may be erroneous, but, in aggregate, the resulting ranking can still be correct. Similarly, we observe that listwise approaches (\ItemListSym) are generally more volatile, possibly due to more difficult calibration or positional biases \citep{han_prototypical_2023, wang_large_2023}.

\section{Discussion}
\label{sec:discussion}

To scrutinize our results, we explore settings with a train-test split, and discuss the interpretability considerations of \probing.

\paragraph{Ranking Direction across Tasks.}

Instead of training our probes on a single ranking task, we train them on a training set of multiple rankings and evaluate on a held-out set. To this end, we use \num{4}-fold cross-validation which allows comparing \probing against supervised probing in a fair setup. This setup is more similar to the experiments in the original \ccs paper \citep{burns_discovering_2023} and thus rests on a similar hypothesis: is there are a more universal \setQuote{ranking direction} in the activations of a language model that holds across ranking tasks? \cref{fig:k-fold} in the appendix presents the results of this k-fold validation experiment. Firstly, our probes identify ranking properties that exist across different ranking tasks. This particularly holds for ranking tasks that resemble each other more closely as in \ScalarAdj or \Reviews. Secondly, \probing does not fall far behind supervised probing. Since this is especially evident for datasets with fewer ranking tasks, we hypothesize that \probing is less likely to overfit and instead exploits general ranking properties.

\paragraph{Interpretability.}

\begin{figure}[t]
 \centering
 \includegraphics[width=1.0\linewidth]{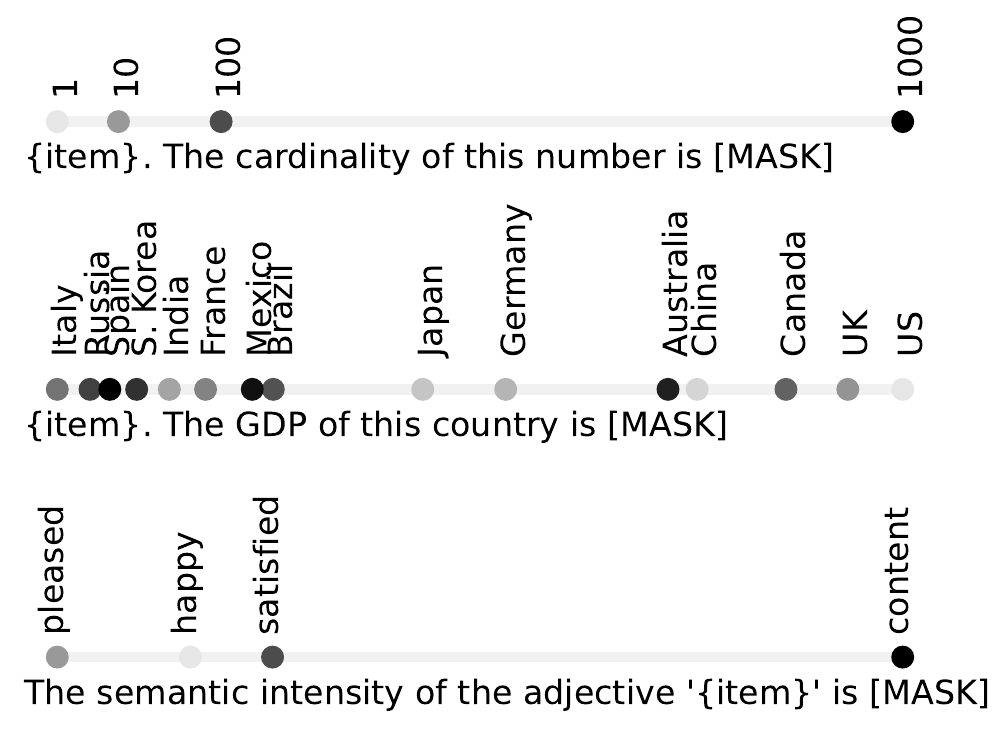} 
\caption{\probing offers interpretability benefits such as the post-hoc analysis of the probe's parameters. The gray scale hue of the individual dots represents the ground truth ranking of the respective items.}
\label{fig:interpret}
\end{figure}

Besides performance, another argument for \probing is control and post-hoc interpretability offered by the parametric probe. In \cref{fig:interpret} for instance, we plot the ranking scores $\score_{n} = \sigmoid(\boldsymbol{\theta}^{T} \boldsymbol{x}_{n} + b)$ for each item predicted by the linear probing model trained with \TripletCCS. This allows us to inspect the distances between items projected onto the latent ranking scale. The predictions and parameters are deterministic opposed to prompt-based generations from stochastic decoding methods. On a more abstract level, we relate multiple language model queries through a surrogate model that projects the language model's outputs to a shared ranking scale.

\section{Related Work}

This paper builds upon \CCS (\ccs) \citep{burns_discovering_2023}, which has inspired multiple other follow-up works: some explore calibrated versions of \ccs \citep{tao_calibrated_2023}, others adapt \ccs to order-invariant, multi-class settings \citep{zancaneli_adapting_2023}, compare different \ccs objective functions \citep{fry_comparing_2023} and elicit inference-time interventions to increase truthfulness \citep{li_inference-time_2023}. 

\citet{farquhar_challenges_2023} raise concerns that \ccs may not \setQuote{discover knowledge}, but instead simply latches onto to most salient features. We argue that our \ccr approach is less affected by this concern as we are mainly focused on achieving good predictive performance in unsupervised ranking tasks by making consistent measurements across multiple prompts. To this end, we test our method against regression-based variants in \cref{sec:listProbing} and evaluate on an unseen held-out set in \cref{sec:discussion}. 

Pairwise and listwise prompting have been explored in different tasks \citep{ma_zero-shot_2023, lee_prompt-based_2023, liusie_zero-shot_2023}, but is most frequently focused on document retrieval \citep{ferraretto_exaranker_2023}. Pairwise (RankNet) \citep{burges_learning_2005} and listwise (ListNet) \citep{cao_learning_2007} ranking approaches have also been compared outside of language model prompting. We additionally explore pointwise prompting \citep{fu_gptscore_2023} and find that, counter-intuitively, pointwise often outperforms listwise prompting. To move beyond prompting, we propose an expansion of the \ccs method to rankings. \ccs and \ccr are conceptually different to \setQuote{contrast consistency} which refers to contrastive data perturbations \citep{gardner_evaluating_2020, zhang_exploring_2023}. They are also different to \setQuote{contrastive decoding} \citep{li_contrastive_2023} which contrasts log-probabilities between an expert and an amateur model. Instead, our \probing approach is strongly influenced by unsupervised ranking \citep{frydenlund_language_2022} and probing of semantic, ordinal axes \citep{gari_soler_bert_2020, li_discovering_2022, stoehr_sentiment_2023, stoehr_ordered_2023}.

\section{Conclusion} 

We analyze the ranking capabilities of language models by comparing pairwise, pointwise and listwise prompting techniques and find that listwise prompting is less computationally expensive, but more susceptible to mistakes. We then propose an unsupervised probing method termed \CCR (\ccr). \ccr learns an affine mapping between a language model's activations and a model-inherent ranking direction. Especially for smaller language models, \ccr outperforms \prompting, is easier to control, less susceptible to prompt design and more interpretable. We see a lot of potential in in-context probing for making consistent measurements with language models.

\section*{Acknowledgments}

This work was completed while the first author, Niklas Stoehr, did a research internship at Bloomberg. We would like to thank Ozan Irsoy, Atharva Tendle, Faner Lin, Ziyun Zhang, Ashim Gupta, Suchin Gururangan, and the entire Bloomberg AI group for valuable feedback on the manuscript. We would like to express special thanks to Kevin Du and Luca Beurer-Kellner from ETH Z{\"u}rich for early-stage discussions.

\section*{Limitations}
\label{sec:limitations}

We methodologically compare pairwise, pointwise and listwise \prompting and \probing approaches as illustrated in \cref{fig:overview}. One may argue that our proposed versions of pointwise and listwise \probing violate this categorization because pointwise \ccr uses a pairwise loss objective. Similarly, the loss objective of listwise \ccr may be listwise, but the prompt type is \ItemSingle. To draw the distinction, we consider prediction time at which the probe trained with \MarginCCS or \TripletCCS outputs a single, thus pointwise, ranking score per item (see \cref{fig:interpret}). Similarly, the probe trained with \OrdRegCCS predicts a full vector of scores for all (listwise) ranks. Yet, we do encourage future work to explore further pointwise and listwise \probing approaches.

The direction-invariance of both \ccs and \ccr poses another potential limitation that may be lifted by future work as further outlined in \cref{sec:ranking_direction}. In particular, for pointwise and listwise prompting, omitting the direction of a desired ranking can hurt performance. The language model may be confused whether to rank the highest or lowest item first, leading the items' corresponding logit scores to cannibalize each other. This weakness of prompting may be interpreted as a strength of \probing however, as it is less prompt-sensitive. An important direction for future work is testing \prompting and \probing in ranking tasks with even larger or instruction-tuned language models. 

Since we do not consider a train-validation-test set split in this work, we refrain from hyperparameter-tuning (e.g., margins, learning rate, sub-batching, probe initialization). However, based on initial prototyping, we see performance boosts for \ccr when tuning these hyperparameters. We envision further boost in \probing performance through more expressive probing models, e.g., non-linear kernels or neural networks. Yet, the admissible number of probe parameters and the requirement to use the same probe for different ranking tasks irrespective of their number of items are limiting factors.

\section*{Impact Statement}

Throughout this work, we evaluate language models in \setQuote{transformative} rather than \setQuote{generative} tasks---we avoid any free-form generation and strongly constrain a model's output to an explicit list of answer candidates. Moreover, the focus of this work lies on mitigating model hallucinations in the context of ranking.
We pursue this goal in two ways: on the one hand, testing a model's parametric ranking-based knowledge may indicate knowledge gaps, outdated information or biases. On the other hand, constraining a model's output in in-context reasoning tasks leads to more consistent and thus more truthful ranking results. All datasets considered in this work are publicly available, but are in English only. We thoroughly checked all licensing terms and adhered to the intended use of the data, We also manually verified that the data do not contain personally identifiable information.

\bibliography{references}
\bibliographystyle{acl_natbib}

\appendix

\section{Appendix}
\label{sec:appendix}

\paragraph{Direction-invariance of \ccs and \ccr.}
\label{sec:ranking_direction}

We limit the scope of this work to direction-invariant rankings: i.e., the ranking $A > B > C$ is considered to be the same as $C > B > A$. This assumption aligns well with the original \CCS (\ccs) method \citep{burns_discovering_2023}. In \ccs, the probe is trained to map statements and their negation to either a \num{0} or \num{1} pole consistently across multiple paired statements. However, it is not defined a priori, which of the two poles corresponds to all truthful and all false statements. We argue that this is even less a shortcoming for \ccr than it is for \ccs. While the meaning of the poles, \setQuote{true} versus \setQuote{false} for \ccs, \setQuote{high rank} versus \setQuote{low rank} for \ccr, needs to by interpreted post-hoc, the ordering of items obtained with \ccr can be directly read off. With \origCCS, the probe predicts the label of a new statement according to
\begin{align}
\score_{i} = \frac{1}{2} \Big(\probe (\itemPlusVec) - \big(1 - \probe (\itemNegVec)\big)\Big)
\end{align}
In the case of \MarginCCS, \TripletCCS and \OrdRegCCS, the probe directly predicts a ranking score $\score_{n}$, because items are represented by individual vectors via the \ItemSingle prompt type.

\paragraph{Bias Terms for \OrdRegCCS.}
\label{sec:bias_terms}

The CORAL model \citep{cao_rank_2020} used in combination with the \OrdRegCCS objective (\cref{sec:listProbing}) comprises $K$ bias terms $b_k$. Since we would like to limit the number of parameters, we parametrize these bias terms via a polynomial function with learnable parameters $\alpha$ and $\beta$. We first cut a $[0,1]$ interval into $K-1$ unit-length pieces with the cut-off points $\{\delta_k\}_1^{K-1}$. We then transform these points through a polynomial function $g_{\alpha,\beta}$ as follows
\begin{align}
\delta_k^{'} = g_{\alpha,\beta} \big( \delta_k^{(a-1)} (1-\delta_k)^{(b-1)} \big)
\end{align}
The function $g$ is parametrized by only two parameters $\alpha$ and $\beta$ similar to the Beta function. As an uninformative prior, we set $\alpha = 1.0$ and $\beta = 1.0$ and optimize the parameters during inference. The transformed cut-off points $\delta_k^{'}$ are further shifted to ensure they are monotonically decreasing and centered around \num{0}. To this end, we first compute the reverse (right-to-left) cumulative sum according to $\delta^{''}_k = \sum_{k=0}^{K-2} \delta^{'}_{K-k}$. Finally, we compute the mean $\bar{\delta^{''}} = \frac{\sum_1^{K-1}\delta^{''}_k}{K-1}$ which we subtract from every transformed $\delta^{''}_k$ to finally obtain $b_k$.

\paragraph{Technical Details.}
\label{sec:technical_details}

In all \probing setups, we dynamically set the batch size to the number of items of a ranking task. For the pairwise approaches, we perform sub-batching with two items at a time. For the approaches based on \ItemSingle, we Z-score normalize all vector representations in a batch. We set the margin $m = 0.2$ and include an additional positive margin term in \TripletCCS to avoid the anchor and positive item to collapse to the same value. We train all supervised and unsupervised probes using the Adam optimizer \citep{kingma_adam_2015} with its default settings for \num{200} epochs. Experiments were run on a MacBook Pro M1 Max (64 Gb) and a NVIDIA TITAN RTX GPU. We publish code and data at \href{https://github.com/niklasstoehr/contrast-consistent-ranking}{github.com/niklasstoehr/contrast-consistent-ranking}.

\begin{figure*}[ht]
 \centering
 \includegraphics[width=1.0\linewidth]{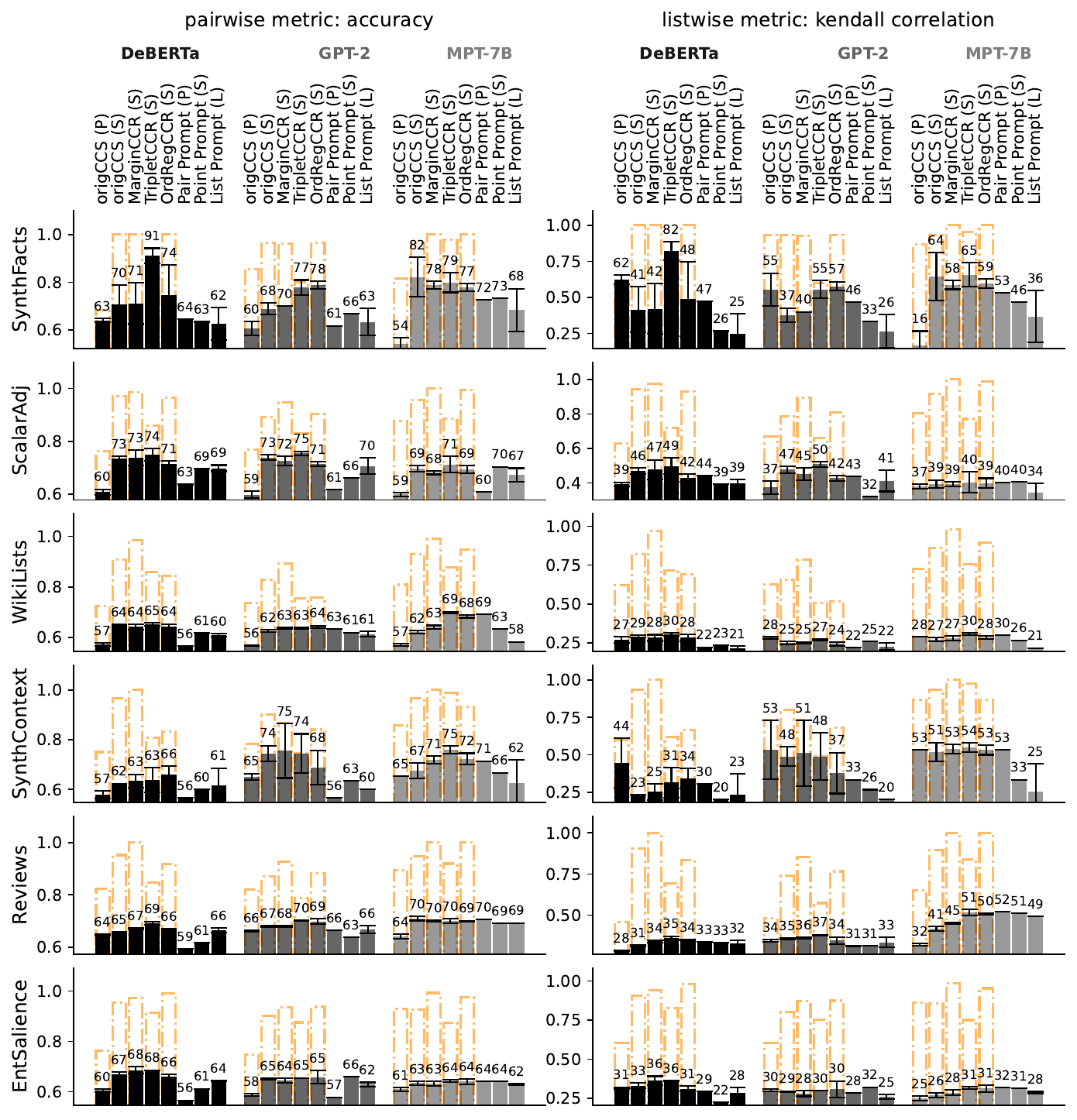} 
\caption{Mean ranking results and standard deviation for all methods and datasets over \num{5} runs.}
\label{fig:per_ranking}
\end{figure*}

\begin{figure*}[ht]
 \centering
 \includegraphics[width=1.0\linewidth]{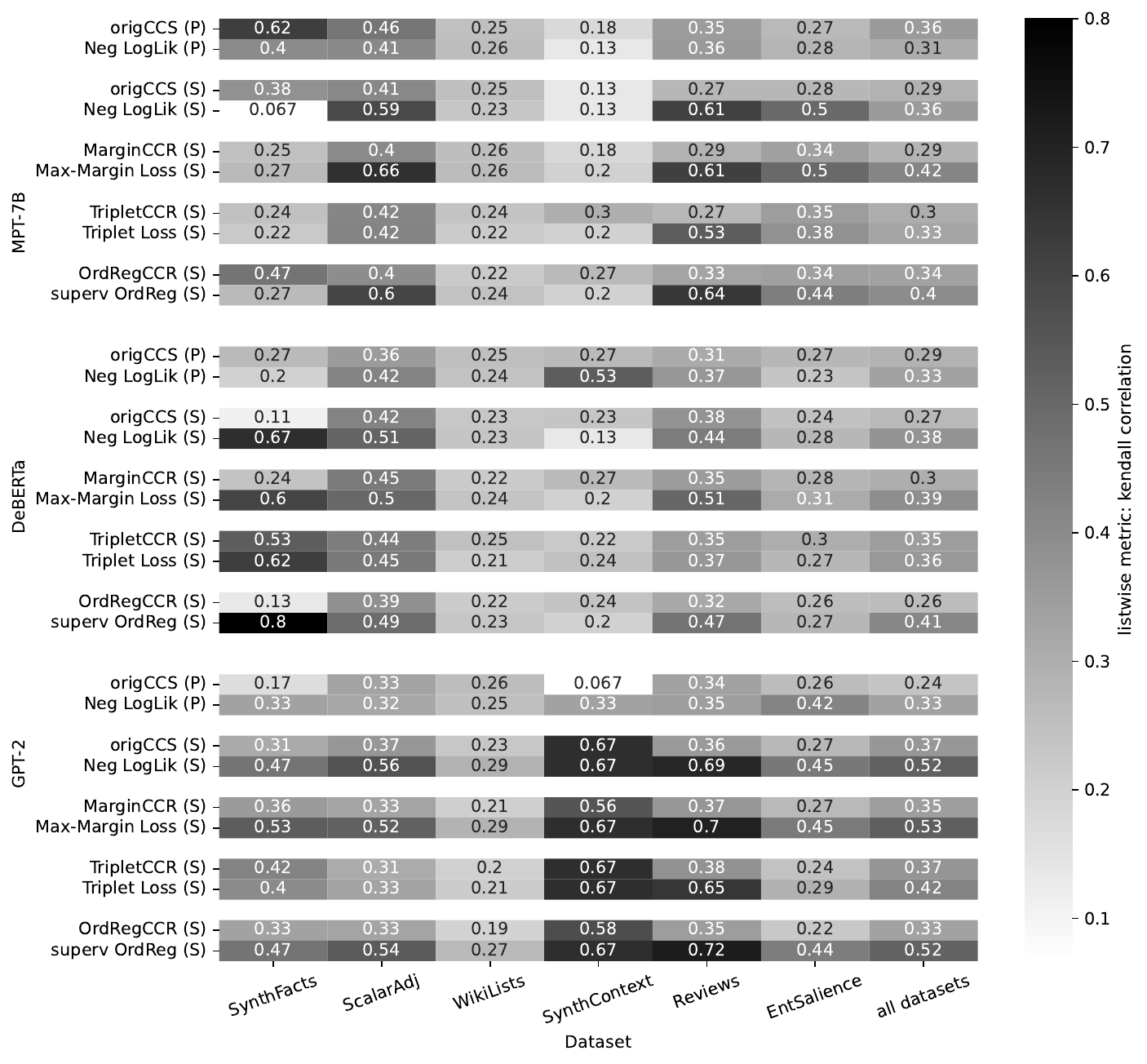} 
\caption{K-fold cross-validation results comparing unsupervised \probing and supervised probing.}
\label{fig:k-fold}
\end{figure*}

\FloatBarrier

\begin{table*}[ht]
\fontsize{9}{9}\selectfont
\centering
\renewcommand{\arraystretch}{1.3} 
\setlength{\tabcolsep}{0.50em} 
\begin{tabular}{l|l}
\textbf{\SynthFacts}          & \textbf{}                                                                                                                                                                                                                                                                                                                                                                                                                                \\ \hline
sentiment of the adjective    & horrible, bad, okay, good, great, awesome                                                                                                                                                                                                                                                                                                                                                                                                \\
cardinality of the number     & 1, 10, 100, 500, 1000, 10000                                                                                                                                                                                                                                                                                                                                                                                                             \\
                              &                                                                                                                                                                                                                                \\
\textbf{\SynthContext}        &                                                                                                                                                                                                                                                                                                                                                                                                                                          \\ \hline
popularity of the color       & \begin{tabular}[c]{@{}l@{}}context: Most students selected blue as their favourite color, followed by red, then yellow. \\ Brown ranked lowest, green second lowest and purple third lowest; \\ items: brown, green, purple, yellow, red, blue\end{tabular}                                                                                                                                                                              \\
wealth of people              & \begin{tabular}[c]{@{}l@{}}context: An owns 100 dollar, Tom owns 50 dollars more and Sam 75 dollars more. Jenny is\\  the richest owning 1000 dollar. Emily and Muhammad are at the lower end owning \\ only 5 dollar and 10 dollars respectively.\\ items: Emily, Muhammad, An, Tom, Sam, Jenny\end{tabular}                                                                                                                                    
\end{tabular}
\caption{Details of our synthetic ranking task datasets \SynthFacts and \SynthContext.}
\label{tab:synth_data_details}
\end{table*}

\begin{table*}[ht]
\fontsize{6.2}{6.2}\selectfont
\centering
\renewcommand{\arraystretch}{1.0} 
\setlength{\tabcolsep}{0.60em} 
\begin{tabular}{ll}
Buildings by volume                                                & \url{https://en.wikipedia.org/wiki/List_of_largest_buildings}                                   \\
Buildings by floor area                                            & \url{https://en.wikipedia.org/wiki/List_of_largest_buildings}                                   \\
Buildings by height                                                & \url{https://en.wikipedia.org/wiki/List_of_tallest_buildings}                                   \\
Airports by passenger traffic                                      & \url{https://en.wikipedia.org/wiki/List_of_busiest_airports_by_passenger_traffic}               \\
Museums by visitors                                                & \url{https://en.wikipedia.org/wiki/List_of_most-visited_museums}                                \\
Tallest church buildings                                           & \url{https://en.wikipedia.org/wiki/List_of_tallest_church_buildings}                            \\
Football stadiums by capacity                                      & \url{https://en.wikipedia.org/wiki/List_of_association_football_stadiums_by_capacity}           \\
Tallest statues                                                    & \url{https://en.wikipedia.org/wiki/List_of_tallest_statues}                                     \\
Architectural Styles                                               & \url{https://en.wikipedia.org/wiki/Timeline_of_architectural_styles}                            \\
Periods in art history                                             & \url{https://en.wikipedia.org/wiki/Periods_in_Western_art_history}                              \\
Plays by Shakespeare by time                                       & \url{https://www.britannica.com/topic/list-of-plays-by-Shakespeare-2069685}                     \\
Operas by Puccini by premiere date                                 & \url{https://en.wikipedia.org/wiki/List_of_compositions_by_Giacomo_Puccini}                     \\
Most expensive paintings sold                                      & \url{https://en.wikipedia.org/wiki/List_of_most_expensive_paintings}                            \\
Planets in the solar system by size                                & \url{https://en.wikipedia.org/wiki/List_of_Solar_System_objects_by_size}                        \\
Planets in the solar system by distance from the Sun               & \url{https://en.wikipedia.org/wiki/Solar_System}                                                \\
Moons of Jupiter by radius                                         & \url{https://en.wikipedia.org/wiki/List_of_Solar_System_objects_by_size}                        \\
Heaviest terrestrial animals                                       & \url{https://en.wikipedia.org/wiki/Largest_and_heaviest_animals}                                \\
Chemical elements by atomic number                                 & \url{https://en.wikipedia.org/wiki/List_of_chemical_elements}                                   \\
Chemicals by boiling point                                         & \url{https://en.wikipedia.org/wiki/Melting_point}                                               \\
Chemicals by melting point (highest to lowest)                     & \url{https://en.wikipedia.org/wiki/Melting_point}                                               \\
Materials by hardness on Mohs scale                                & \url{https://en.wikipedia.org/wiki/Mohs_scale}                                                  \\
Countries by population                                            & \url{https://en.wikipedia.org/wiki/List_of_countries_and_dependencies_by_population}            \\
US Counties by population                                          & \url{https://en.wikipedia.org/wiki/List_of_the_most_populous_counties_in_the_United_States}     \\
Capital cities by elevation                                        & \url{https://en.wikipedia.org/wiki/List_of_capital_cities_by_elevation}                         \\
Metropolitan areas by size                                         & \url{https://en.wikipedia.org/wiki/List_of_largest_cities}                                      \\
Religions by followers                                             & \url{https://en.wikipedia.org/wiki/List_of_religious_populations}                               \\
Ethnic groups by size in the US                                    & \url{https://en.wikipedia.org/wiki/Race_and_ethnicity_in_the_United_States}                     \\
Countries by unemployment rate according to OECD                   & \url{https://en.wikipedia.org/wiki/List_of_countries_by_unemployment_rate}                      \\
Oil producing countries                                            & \url{https://en.wikipedia.org/wiki/List_of_countries_by_oil_production}                         \\
GDP per capita                                                     & \url{https://en.wikipedia.org/wiki/List_of_countries_by_GDP_(nominal)_per_capita}               \\
Wine producing countries                                           & \url{https://en.wikipedia.org/wiki/List_of_wine-producing_regions}                              \\
Largest power stations                                             & \url{https://en.wikipedia.org/wiki/List_of_largest_power_stations}                              \\
Tourists for city                                                  & \url{https://en.wikipedia.org/wiki/List_of_cities_by_international_visitors}                    \\
Total energy from solar sources by country                         & \url{https://en.wikipedia.org/wiki/Solar_power_by_country}                                      \\
Solar capacity as share of total energy consumption by country     & \url{https://en.wikipedia.org/wiki/Solar_power_by_country}                                      \\
Countries by size                                                  & \url{https://en.wikipedia.org/wiki/List_of_countries_and_dependencies_by_area}                  \\
US Counties by area                                                & \url{https://en.wikipedia.org/wiki/List_of_the_largest_counties_in_the_United_States_by_area}   \\
US States by area                                                  & \url{https://en.wikipedia.org/wiki/List_of_U.S._states_and_territories_by_area}                 \\
Lakes by surface                                                   & \url{https://en.wikipedia.org/wiki/List_of_lakes_by_area}                                       \\
Lakes by depth                                                     & \url{https://en.wikipedia.org/wiki/List_of_lakes_by_depth}                                      \\
Rivers by length                                                   & \url{https://en.wikipedia.org/wiki/List_of_rivers_by_length}                                    \\
Mountains by height                                                & \url{https://en.wikipedia.org/wiki/List_of_highest_mountains_on_Earth}                          \\
Islands by surface area                                            & \url{https://en.wikipedia.org/wiki/List_of_islands_by_area}                                     \\
Volcanoes by height                                                & \url{https://en.wikipedia.org/wiki/List_of_volcanoes_by_elevation}                              \\
Waterfalls by height                                               & \url{https://en.wikipedia.org/wiki/List_of_waterfalls_by_height}                                \\
Caves by depth                                                     & \url{https://en.wikipedia.org/wiki/List_of_deepest_caves}                                       \\
Oceans by area                                                     & \url{https://en.wikipedia.org/wiki/Ocean}                                                       \\
Oceans by coastline                                                & \url{https://en.wikipedia.org/wiki/Ocean}                                                       \\
Oceans by average depth                                            & \url{https://en.wikipedia.org/wiki/Ocean}                                                       \\
Deserts by area                                                    & \url{https://en.wikipedia.org/wiki/List_of_deserts_by_area}                                     \\
Oceanic trenches                                                   & \url{https://en.wikipedia.org/wiki/Oceanic_trench#Deepest_oceanic_trenches}                     \\
Countries by area                                                  & \url{https://en.wikipedia.org/wiki/List_of_countries_and_dependencies_by_area}                  \\
Canyons by depth                                                   & \url{https://www.worldatlas.com/canyons/10-deepest-canyons-in-the-world.html}                   \\
Oldest reigning monarchs                                           & \url{https://en.wikipedia.org/wiki/List_of_longest-reigning_monarchs}                           \\
Presidents of the US                                               & \url{https://en.wikipedia.org/wiki/List_of_presidents_of_the_United_States}                     \\
Sultans of the Ottoman Empire                                      & \url{https://en.wikipedia.org/wiki/List_of_sultans_of_the_Ottoman_Empire}                       \\
Emperors of Rome                                                   & \url{https://en.wikipedia.org/wiki/List_of_Roman_emperors}                                      \\
Kings of Rome                                                      & \url{https://en.wikipedia.org/wiki/King_of_Rome}                                                \\
List of time periods in history                                    & \url{https://en.wikipedia.org/wiki/List_of_time_periods}                                        \\
Platonic solids by number of faces                                 & \url{https://en.wikipedia.org/wiki/Platonic_solid}                                              \\
Best selling artists by albums                                     & \url{https://en.wikipedia.org/wiki/List_of_best-selling_music_artists}                          \\
Songs with most weeks at number one on the Billboard Hot 100       & \url{https://en.wikipedia.org/wiki/List_of_Billboard_Hot_100_chart_achievements_and_milestones} \\
Football teams by UEFA Champions League trophies                   & \url{https://en.wikipedia.org/wiki/List_of_European_Cup_and_UEFA_Champions_League_finals}       \\
Most Ballon d'Or Trophies                                          & \url{https://en.wikipedia.org/wiki/Ballon_d%27Or}                                               \\
Countries with the most FIFA World Cup trophies                    & \url{https://en.wikipedia.org/wiki/FIFA_World_Cup}                                              \\
Men's tennis players with the most grand slams won in the open era & \url{https://en.wikipedia.org/wiki/List_of_Grand_Slam_men%27s_singles_champions}                \\
Olympic summer games host cities by year                           & \url{https://en.wikipedia.org/wiki/List_of_Olympic_Games_host_cities}                           \\
List of Dutch football champions by number of titles               & \url{https://en.wikipedia.org/wiki/List_of_Dutch_football_champions}                            \\
List of Romanian football cup winners by number of titles          & \url{https://en.wikipedia.org/wiki/Cupa_Rom%C3%A2niei}                                         
\end{tabular}
\caption{Ranking tasks (mostly extracted from Wikipedia) and curated for our \WikiLists dataset.}
\label{tab:wiki_data}
\end{table*}

\end{document}